\documentclass[nhess, manuscript]{copernicus}
\usepackage{scalerel}

\usepackage{booktabs}
\usepackage{fixltx2e}
\begin{document}
\nolinenumbers
%\input{stucke_nhess-concordance}
%\setkeys{Gin}{width=8.3cm}
%\input{stucke_nhess-concordance}
\title{Upward lightning at wind turbines: Risk assessment from larger-scale meteorology}

% \Author[affil]{given_name}{surname}

\Author[1,2]{Isabell}{Stucke}
\Author[1]{Achim}{Zeileis}
\Author[2]{Georg J.}{Mayr}
\Author[1]{Thorsten}{Simon}
\Author[3]{Gerhard}{Diendorfer}
\Author[3]{Wolfgang}{Schulz}
\Author[3]{Hannes}{Pichler}
\Author[1,2]{Deborah}{Morgenstern}

\affil[1]{Department of Statistics, University of Innsbruck, Austria}
\affil[2]{Department of Atmospheric and Cryospheric Sciences, University of Innsbruck, Austria}
\affil[3]{OVE Service GmbH, Dept. ALDIS (Austrian Lightning Detection $\&$ Information System), Vienna, Austria}

%% The [] brackets identify the author with the corresponding affiliation. 1, 2, 3, etc. should be inserted.

%% If an author is deceased, please mark the respective author name(s) with a dagger, e.g. "\Author[2,$\dag$]{Anton}{Smith}", and add a further "\affil[$\dag$]{deceased, 1 July 2019}".

%% If authors contributed equally, please mark the respective author names with an asterisk, e.g. "\Author[2,*]{Anton}{Smith}" and "\Author[3,*]{Bradley}{Miller}" and add a further affiliation: "\affil[*]{These authors contributed equally to this work.}".

\correspondence{Isabell Stucke (isabell.stucke@uibk.ac.at)}

\runningtitle{Upward lightning at wind turbines: Risk assessment from larger-scale meteorology}

\runningauthor{Stucke et al.}

\received{}
\pubdiscuss{} %% only important for two-stage journals
\revised{}
\accepted{}
\published{}

%% These dates will be inserted by Copernicus Publications during the typesetting process.

\firstpage{1}

\maketitle

\begin{abstract}
Upward lightning has become an increasingly important threat to wind turbines as ever more of them are being installed for renewably producing electricity. The taller the wind turbine the higher the risk that the type of lightning striking the man-made structure is upward lightning. Upward lightning can be much more destructive than downward lightning due to its long lasting initial continuous current leading to a large charge transfer within the lightning discharge process. Current standards for the risk assessment of lightning at wind turbines mainly take the summer lightning activity into account, which is inferred from LLS.
Ground truth lightning current measurements reveal that less than 50~\% of upward lightning might be detected by lightning location systems (LLS). This leads to a large underestimation of the proportion of LLS-non-detectable upward lightning at wind turbines, which is the dominant lightning type in the cold season.
This study aims to assess the risk of LLS-detectable and LLS-non-detectable upward lightning at wind turbines using direct upward lightning measurements at the Gaisberg Tower (Austria) and S\"antis Tower (Switzerland). Direct upward lightning observations are  linked to meteorological reanalysis data and joined by random forests, a powerful machine learning technique. The meteorological drivers for the non-/occurrence of LLS-detectable and LLS-non-detectable upward lightning, respectively, are found from the random forest models trained at the towers and have large predictive skill on independent data. In a second step the results from the tower-trained models are extended to a larger study domain (Central and Northern Germany). The tower-trained models for LLS-detectable lightning is independently verified at wind turbine locations in that domain and found to reliably diagnose that type of upward lightning. Risk maps based on case study events show that high diagnosed probabilities in the study domain coincide with actual upward lightning events. This lends credence to the transfer of the model for all upward lightning types, which increases both the risk and the affected areas.

\end{abstract}

%\copyrightstatement{TEXT} %% This section is optional and can be used for copyright transfers.

\section{Introduction}\label{sec:introduction}

The growing importance to produce renewable energy has recently led to a notable increase in the number of wind turbines \citep[e.g.,][]{pineda2018}. Since those structures are commonly taller than $100$~m, the initiation of upward lightning (UL) propagating from the tall structure towards the clouds is facilitated \citep{berger1967}. A tall structure is more prone to experience UL as it is exposed to a stronger electrical field in comparison to the ground. Structures shorter than $100$~m mainly experience downward lightning (DL) with leaders propagating from the clouds towards the earth surface \citep[e.g.,][]{rakov2003}.

As wind turbines are getting taller, UL is the major weather-related cause of severe damages to them \citep[e.g.,][]{rachidi2008, montana2016,pineda2018,matsui2020,zhang2020}. It can be much more destructive than DL due to its initial continuous current (ICC) lasting approximately ten times longer than the current of  DL. 
Ground truth lightning current measurements at the specially instrumented tower on top of the Gaisberg mountain (Austria, Salzburg) reveal that more than $50$~\% of UL is not detected by conventional lightning location systems (LLS). The reason is that the LLS cannot detect a particular subtype of UL having only an ICC \citep{diendorfer2015,march2016}.
Even though towers exist providing ground truth lightning current data for LLS-detectable UL (UL-LLS) such as the S\"antis Tower in Switzerland, the Gaisberg Tower is the only instrumented tower in Europe providing the full information on the occurrence of both UL-LLS and LLS-non-detectable UL (UL-noLLS). %This highlights the impossibility to infer the exact risk of UL at wind turbines from LLS alone. 

Standards for lightning protection of wind turbines \citep[e.g.,][]{lightningstandards} crucially underestimate the occurrence of UL at wind turbines since they currently rely only on three factors: The height of the wind turbine, the local annual flash density derived from LLS and an environmental term involving factors like terrain complexity or altitude \citep{rachidi2008,pineda2018}. Lightning activity in summer clearly dominates the annual local flash density due to large amounts of  DL caused by deep convection. 
%The occurrence of UL is currently only considered by adding the environmental term which leads to a significant underestimation of UL.
However, UL is expected to be the dominant lightning type at wind turbines with a tendency to be even more important in the colder season \citep{diendorfer2020,rachidi2008}. Further the risk assessment standards cannot account for UL-noLLS, but can only account for UL-LLS given that a tall structure is present. 
%However, the risk assessment standards only account for UL in the warm season and highly underestimate UL in the colder season \citep[e.g.,][]{pineda2018}. Further the risk assessment standards only account for LLS-detectable UL and neglect LLS-non-detectable UL. 

%The increasing need for a proper risk assessment of UL at wind turbines motivates to introduce the objective of this study. 
The major objective of this study is to assess the risk of UL-LLS and UL-noLLS at wind turbines over a larger domain.
Only at very few points the actual occurrence of UL can be analyzed based on direct measurements. Even though LLS networks exist which might allow to analyze UL-LLS at tall structures, the lightning current measurements show that a significant proportion is missed. 
Being aware that conventional LLS cannot assess the full risk of UL at wind turbines, this study uses a new approach. 

It uses machine learning techniques linking the occurrence of UL to the larger-scale meteorological setting. 
The occurrence of UL can only be provided by ground truth lightning current measurements. These are the basis to build and train the statistical models used to eventually assess the risk of UL over a whole study domain. 
Specifically, this study employs conditional inference random forests \citep{hothorn2015}, which account for highly nonlinear and complex interactions between the incidence of UL to the tall structures and the atmosphere.
The achievement of the major objective requires several steps.

From lightning current measurement data at two instrumented towers in Austria (Gaisberg Tower) and Switzerland (S\"antis Tower) two models are constructed: One for UL-LLS and one for UL-LLS + UL-noLLS. These shall first find whether there is a relationship between larger-scale meteorological variables and the occurrence of UL and second demonstrate how well larger-scale meteorology can serve as a diagnostic tool to infer the occurrence of UL. 

The benefit of the availability of UL-LLS data helps to verify whether the results from the instrumented towers are transferable. The idea is to extract wind turbine locations within the study domain and identify all lightning strikes to them from the colder season (ONDJFMA) using LLS data. Succeeding in reliably diagnosing UL-LLS from larger-scale meteorology in combination with UL ground truth lightning current measurements provides a stronger reliability of the results when in a final step the risk of UL-noLLS, which cannot be verified using LLS data, is assessed.

The following sections are organized as follows.
Section \ref{sec:data} introduces the two instrumented towers providing the necessary ground truth data for this study. The first one is the Gaisberg Tower providing both UL-LLS and UL-noLLS and the second one is the S\"antis Tower providing only UL-LLS. Further this section introduces the identification of lightning at wind turbines in the study domain as well as the meteorological data used. 

Section~\ref{sec:methods} summarizes the procedures and major findings from the two instrumented towers. Section~\ref{sec:methods1} describes the basic principle of the construction of a random forest model. Section~\ref{sec:methods2} presents the performance of the models at the instrumented towers. Further, the most important larger-scale meteorological variables are introduced which lead to a higher risk of UL (Sect.~\ref{sec:methods3}).

Then, Sect.~\ref{sec:results} presents the results extending the models from the instrumented towers to the larger study domain to find regions with a higher risk to experience UL. Section~\ref{sec:results1} diagnoses UL-LLS at wind turbines and presents case studies. Then, in Sect.~\ref{sec:results2} the risk of UL-LLS and UL-LLS + UL-noLLS at wind turbines is illustrated and discussed using the whole period of consideration.

% First, they diagnose the probability of UL on selected case study days (Sect.~\ref{sec:casestudies}). Second, they diagnose the overall risk of UL for certain regions within the study domain (Sect.~\ref{sec:potential}).
%Diagnostic performance of the models on the larger domain is assessed by comparing the diagnosed probabilities of UL with actual lightning observations at wind turbines. These are identified in the colder seasons between $2018$ and $2020$ within the larger study domain. Wind turbine locations are extracted using OpenStreetMap data and remotely measured lightning data provided by the European Cooperation for Lightning Detection (EUCLID) provides information when a wind turbine experienced lightning. 

Section~\ref{sec:conclusions} concludes and summarizes the most important findings.

\section{Data}\label{sec:data} 
This study combines five different data sources: UL data measured directly at the Gaisberg Tower in Austria \citep{diendorfer2009} and at the S\"antis Tower in Switzerland \citep{romero2012}; LLS data measured remotely by the European Cooperation for Lightning Detection \citep[EUCLID,][]{schulz2016}; larger-scale meteorological variables from the reanalysis database ERA5 \citep{hersbach2020}; wind turbine locations identified using the OpenStreetMap database.

\subsection{Direct UL measurements at instrumented towers}\label{sec:lightningdata}

Figure \ref{fig:windmills_topomap} shows two of the very few instrumented towers for the direct measurement of currents from UL. These are the Gaisberg Tower (1\,288 m amsl, $47 ^\circ 48 '$ N, $13 ^\circ 60'$ E) and the S\"antis Tower (2\,502 m amsl, $47 ^\circ 14 '$ N, $9 ^\circ 20'$ E).
Lightning at the instrumented towers is almost exclusively UL. 
Gaisberg Tower recorded in total $819$ UL events between $2000$ and $2015$. S\"antis Tower recorded $692$ UL events between $2010$ and $2017$. 

A sensitive shunt type sensor at Gaisberg allows to measure all types of upward flashes regardless of the current waveform, i.e., UL-LLS and UL-noLLS. However, inductive sensors employed at S\"antis cannot measure upward flashes with only an ICC \citep[approximately $50$~\%, ][]{diendorfer2015}. 

Direct UL current measurements are the crucial prerequisite to construct the random forest models, which are extended to the larger study domain after being trained on the tower data. The combination of data from both towers allows to construct the two types of models, that shall diagnose UL-LLS and both UL-LLS + UL-noLLS.

\subsection{UL-LLS at wind turbines and study domain}\label{sec:lightningdata}

Remotely detected lightning data by the LLS EUCLID and wind turbine locations derived from OpenStreetMap serve as verification of the statistical models assessing the risk of UL-LLS for the selected study domain.

\begin{figure}[H]
\begin{center}
 \includegraphics[width=0.9\textwidth]{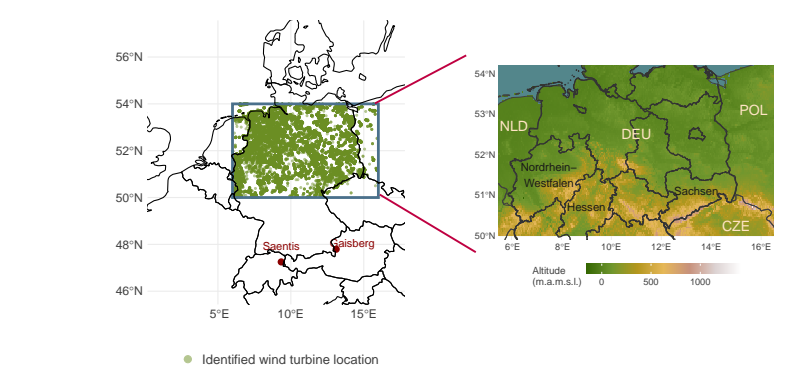}
  \caption{Geographic overview of the instrumented tower locations (Gaisberg and S\"antis) as well as the study domain (box). Green dots are manually identified wind turbine locations based on  © OpenStreetMap $2020$. Right: topographic map of study domain showing altitude above mean sea level. Data taken from Shuttle Radar Topography Mission \citep{farr2000}.}
  \label{fig:windmills_topomap}
 \end{center}
\end{figure}

Within the study domain of 50°N--54°N and 6°E--16°E, $27~814$ wind turbines have been installed by the end of $2020$ (Fig.~\ref{fig:windmills_topomap}). Having extracted the exact locations of these wind turbines, lightning strikes within a 0.003° circular area (approximately within 300~m radius) detected by EUCLID are identified and assumed as UL. EUCLID measures DL with a high flash detection efficiency of more than \unit{90\,\%} \citep{schulz2016}. As mentioned, UL might be detected less efficiently \citep[< \unit{50\,\%}][]{diendorfer2015}. %Therefore, lightning provided by EUCLID may be considered as the lower limit of actually occurred lightning at wind turbines.

%TODO: how to cite IEC 61400-24:2019 Wind energy generation systems - Part 24: Lightning protection

Due to its destructive potential and its severe underestimation in the current lightning protection standards, UL shall be explicitly accounted for investigating the risk of UL at wind turbines in the study domain. The tower-trained models are based on UL data throughout the year. However, as UL is dominant in the colder season compared to DL, only the months from October to April, starting from October $2018$ until December $2020$ are considered in the verification part of the study. Further, since DL is dominant in the warmer season, the extraction of lightning strikes to wind turbines would possibly lead to ambiguity in the identification of DL or UL when considering the whole year. 
% TODO: shorten
%\cite{diendorfer2020} statistically analyzed upward lightning at wind turbines for the years 2017/2018 and 2018/2019 (45°N -- 50°N and 0°W -- 30°E). They showed that $10$\% of the wind turbines experienced at least one lightning strike within the cold season emphasizing to account for the risk of upward lightning in future protection standards due to this comparably high probability.
%\cite{diendorfer2020} further noted an increase of the probability of lightning striking wind turbines with increasing altitudes of the wind turbine locations in this study domain.

\subsection{Meteorological data}\label{sec:ERA5data}
ERA5 provides hourly reanalysis of the state of the atmosphere. It has a resolution of $31$~km horizontally ( grid cell size of 0.25~°~x~0.25~° ) and $137$ levels vertically.
This study uses 35 directly available and derived variables at the surface, on model levels and integrated vertically. These reflect variables relevant for cloud electrification, lightning and thunderstorms  \citep{morgenstern2022}. A full list of variables can be found in the Appendix \ref{sec:appendix}. 
Data are spatially and temporally bilinearly interpolated to each Gaisberg and S\"antis Tower UL observation as well as to each grid cell within the study domain in the verification part of this study. 

\section{Methodological procedures and findings from the instrumented towers}\label{sec:methods}
This section provides the required background information on the basic methods as well as important outcomes from the analysis at the instrumented Gaisberg Tower and S\"antis Tower. Three different aspects shall be covered in the following: First the principle how the basic model, i.e., a random forest, is constructed. Second, the performance of the models and third, which variables are most important to identify favorable conditions for UL to occur or not.

\subsection{Construction and verification of the tower-trained random forests}\label{sec:methods1} %TODO hier schreiben dass schon Ergebnisse sind!!!

A machine learning technique, which has been recently widely adopted in various scientific fields, is used to link larger-scale meteorology and the occurrence of UL at the instrumented towers. Random forests are highly flexible and able to handle nonlinear effects capturing complex interactions with respect to the stated modeling problem \citep{strobl2009}.

%\paragraph*{Principle of random forest models} %TODO: titel ueberdenken
The occurrence versus the non-occurrence of UL is a binary classification problem which is tackled using $35$ larger-scale meteorological variables (predictors). Each meteorological predictor is linked to a situation with or without UL at the Gaisberg or S\"antis Tower using a random forest.
A random forest combines predictions from several decision trees, learned on randomly chosen subsamples of the input data.

Specifically, the trees in the random forest are constructed by capturing the association between the binary response and each of the predictor variables using permutation tests (also known as conditional inference, see  \cite{strasserweber1999}). The idea is that, in each step of the recursive tree construction, the one predictor variable is selected which has the highest (most significant) association with the response variable. Then, the dataset is split with respect to this predictor variable in order to separate the different response classes as well as possible. Splitting is repeated recursively in each of the subsets of the data until a certain stopping criterion (e.g., regarding significance or subsample size) is met. The forest combines $500$ of such trees, where each tree is learned on randomly subsampled two-thirds of the full data set and only considering six randomly selected predictors in each split. Finally, the random forest averages the predictions from the ensemble of trees, which stabilizes and enhances the predictive performance. See \cite{hothorn2006} and \cite{hothorn2015} for more details on the algorithm and implementation.

To validate the resulting models, the input data are split into training and testing data samples. On the training data, the models are trained and on the unseen testing data the diagnostic ability is assessed. 
Leave-one-out cross-validation is used for validating the models for UL-LLS and UL-LLS + UL-noLLS. The first model for UL-LLS uses both S\"antis data and Gaisberg data to increase the size of the training data. The particular flash type that cannot be detected at the S\"antis Tower is left out from the Gaisberg data during the training procedure to ensure consistency. The second model for UL-LLS + UL-noLLS uses only Gaisberg data, as only the Gaisberg Tower provides full information on all subtypes of UL.

Between 2000 and 2015, the Gaisberg Tower experienced $247$ unique days with UL events. Between 2010 and 2017, the S\"antis Tower experienced $186$ unique days. Combining UL days from both towers yields $406$ unique days with UL.
Each training input data leaves out one of the $247$ (406) days with UL to use it as test data. This is repeated until each of the $247$ (406) days has been left out once for training the random forest models. This results in $247$ (406) different models trained on equal numbers of situations with and without UL.

The input model response (i.e., did UL occur or not) is sampled such that the two classes are balanced, i.e., situations with and without UL are present with equal proportions.
Assessing the models' performance, the models diagnose the conditional probability on data not considered during training the models, i.e., on the respective day left out. We call the probability conditional due to the balanced model response setup.
To diagnose the conditional probability of UL on days without UL as well, days without UL from each season are randomly sampled between $2000$ and $2017$. 
High diagnostic ability relates to high probabilities whenever UL occurred at Gaisberg or S\"antis in the particular situation (i.e., a high true positive rate) and low probabilities whenever no UL occurred (i.e., a low false positive rate).

%\section{Procedures and Results}

\subsection{Performance of the tower-trained random forests}\label{sec:methods2}
 
%TODO Saentis model only rausschmeissen!!

\begin{figure}
\begin{center}
 \includegraphics[width=0.4\textwidth]{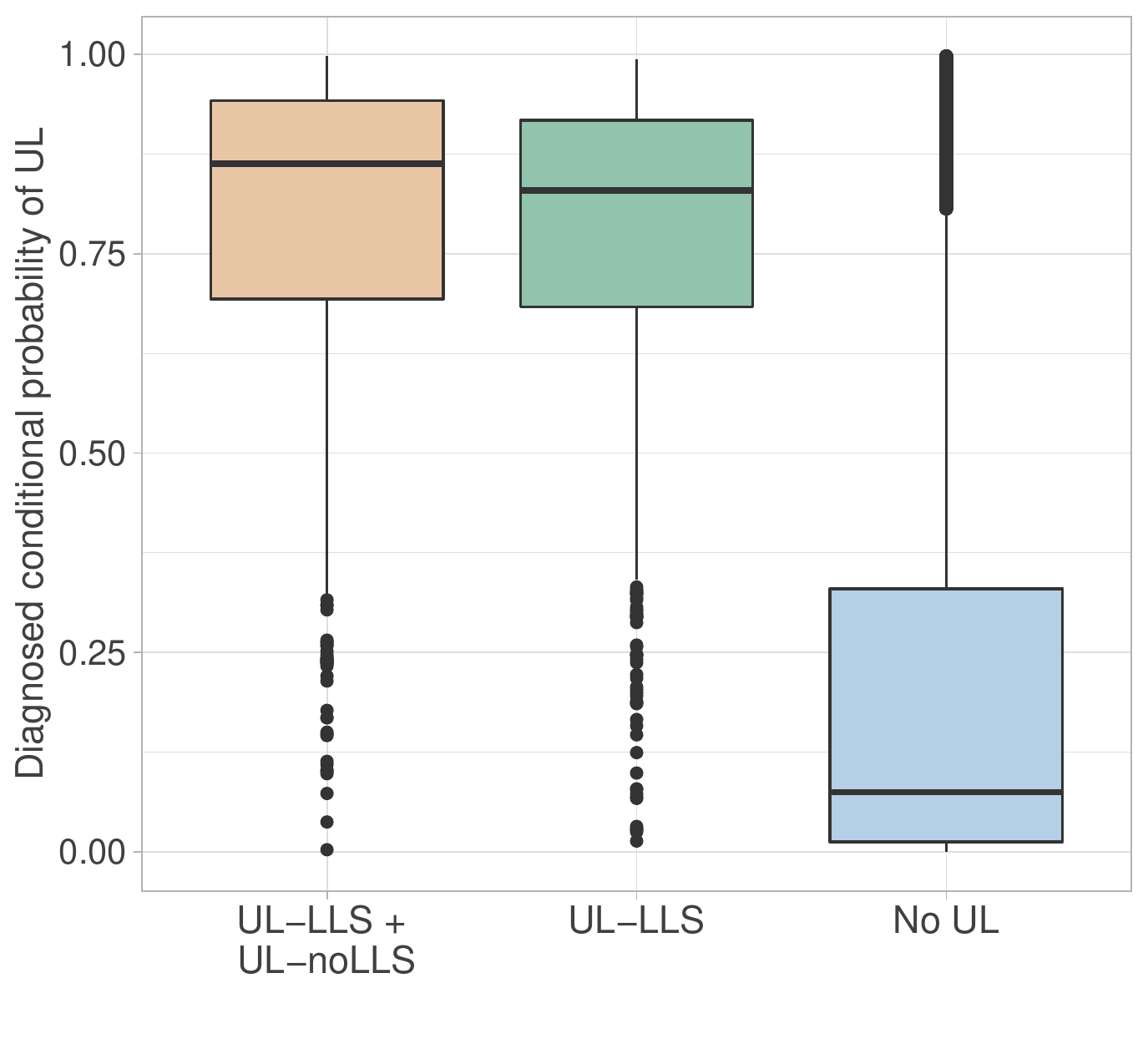}
  \caption{Distributions of diagnosed conditional probabilities in situations with or without UL events. Left: conditional UL probability given that UL was observed in the particular minute (true positive) based on Gaisberg data including all subtypes of UL. Center: conditional UL probability given that UL was observed in the particular minute based on Gaisberg and S\"antis data combined.
Right: conditional UL probability on randomly sampled days without UL events (false positive).}
  \label{fig:diagnostic_abilities}
 \end{center}
\end{figure}

The tower-trained random forest models can reliably diagnose both UL-LLS and UL-LLS + UL-noLLS when validated on unseen withheld data from the towers. 
Figure \ref{fig:diagnostic_abilities} summarizes the cross-validated diagnostic ability according to the random forests for UL-LLS + UL-noLLS (Gaisberg) and UL-LLS (Gaisberg + S\"antis). Both model ensembles show a similarly good diagnostic performance. The diagnosed median conditional probabilities are about $0.8$ given that UL was observed in the respective situation (minute). This indicates a high true positive rate. Similarly, for situations without lightning (right), the conditional probabilities are low indicating a low false positive rate. %Diagnostic ability is highest in winter and in the transition seasons reaching values larger than $0.9$ (not shown here %TODO: zitat iclp).

That the random forest including UL-noLLS has the highest diagnostic ability demonstrates that the proportion which cannot be detected by conventional LLS can be indeed reliably diagnosed by larger-scale meteorology alone. This supports the idea to also investigate the risk for unverifiable UL-noLLS and not only for UL-LLS.  
%The satisfying diagnostic performance using the point data from the instrumented towers provides the basis for an extensive risk assessment of UL over a larger domain.
%Further to assess the models' ability to reliably separate situations with UL from situations without UL from larger-scale meteorological variables alone, the area under the receiver operating characteristic curve is assessed (\citep[AUC,][]{wilks}. For all combinations, the AUC is around $0.8$ and higher, whereby $1$ indicates a perfect separability and $0.5$ indicates no explanatory power. 

%\section{Further Procedures and Results}

\subsection{Meteorological drivers for UL-LLS at the instrumented towers}\label{sec:methods3}

\begin{figure}
\begin{center}
\includegraphics[width=0.5\textwidth]{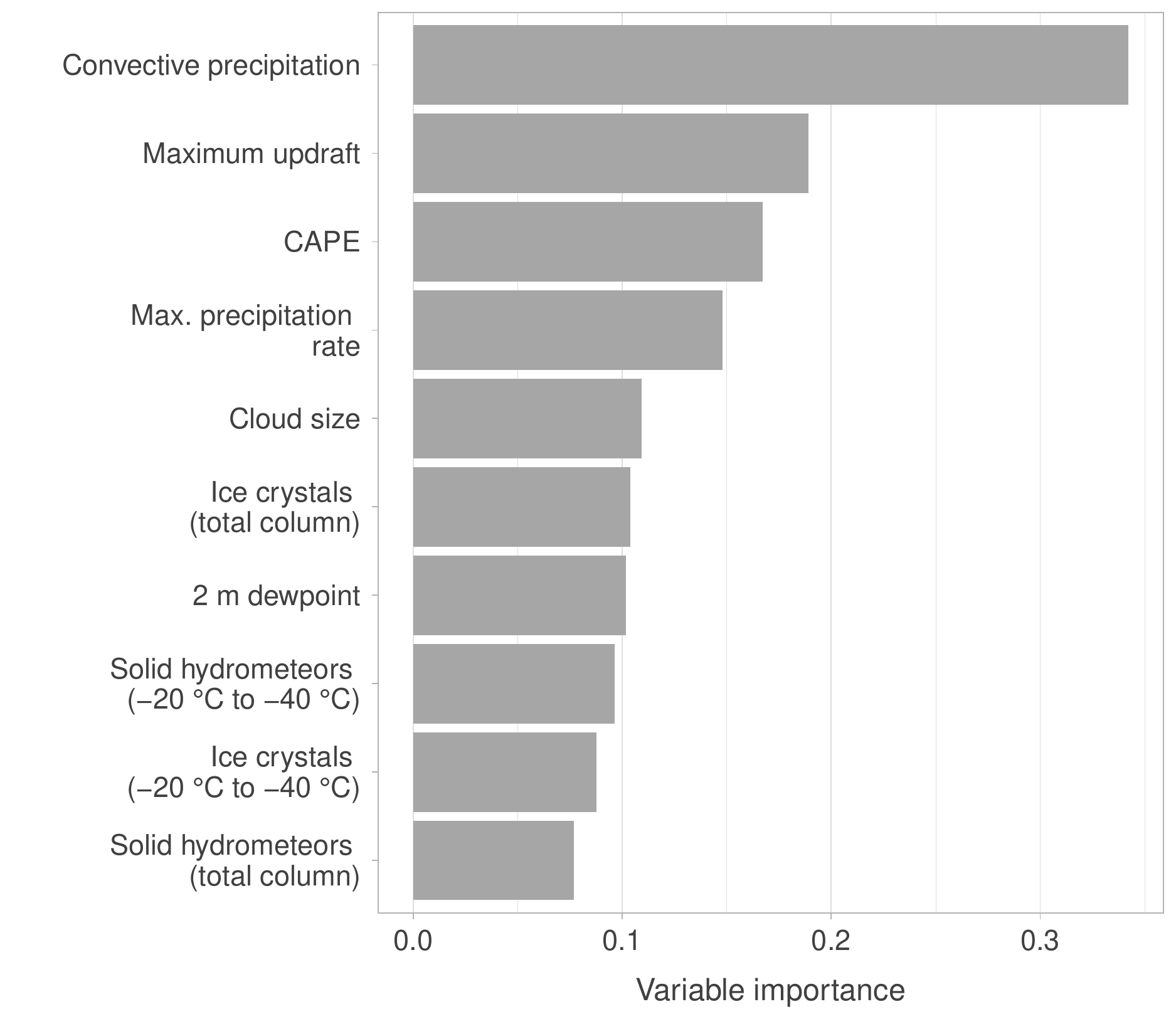}
\caption{Median permutation variable importance according to $100$ different random forests based on balanced proportions of situations with and without UL at the Gaisberg and S\"antis Tower.}
\label{fig:varimp}
\end{center}
\end{figure}

Random forests allow to assess the influence of individual variables on the models' diagnostic performance. This is done by computing the so-called permutation variable importance. The idea is to break up the relationship between the response variable and one predictor variable by neglecting its information when assessing the models' diagnostic performance. Neglecting the information of one predictor variable is done by permutation, i.e., randomly shuffling its values and then assessing how much the diagnostic performance decreases.
Figure~\ref{fig:varimp} visualizes the computed median permutation variable importance according to $100$ different random forest models for UL-LLS. Each of the $100$ models is based on a balanced proportion of situations with UL and randomly chosen situations without UL. Results for UL-LLS and UL-LLS + UL-noLLS models are very similar. %TODO variable importance of LLs non detectable + LLS-detectable

Convective precipitation has the largest influence on the occurrence of UL according to the random forests based on direct observations from the Gaisberg and the S\"antis Tower (Fig.~\ref{fig:varimp}). Neglecting the information of this driver variable reduces the diagnostic performance most. The second and third most important variables are the maximum updraft velocity and convective available potential energy (CAPE), respectively. 
Statistically summarizing the three most important variables shows that CAPE is both at the S\"antis Tower and at the Gaisberg Tower rather low, when UL occurs (median value of $68$~J~kg$^{-1}$). Convective precipitation comes with a median value of $3.8$~mm and maximum vertical updraft velocity with a median of $-~1.5$~m~s$^{-1}$. All values are larger in magnitude than on "average" when considering every single hour in the considered time range. However, in comparison to situations with deep convection, the order of magnitude is not exceptionally high as may be observed with deep convection in which particularly the CAPE values are commonly higher than $500$~J~kg$^{-1}$. An important reason for this might be that at the instrumented towers, UL occurs approximately equally distributed throughout the year whereas intense thunderstorms with deep convection and high CAPE values occur mainly in the summer season. Further this might suggest that for UL to occur, requires a combination of many different processes interacting to form favorable conditions for UL which might be even more complex than providing favorable conditions for deep convection. 

Other important variables are the maximum precipitation rate, the vertical size of the thundercloud, the amount of ice crystals and solid hydrometeors as well as the $2$~m dewpoint temperature are influential.

%\begin{figure}
%\begin{center}
%\includegraphics[width=0.5\textwidth]{varimp}
%\caption{}
%\label{fig:varimp}
%\end{center}
%\end{figure}

\section{UL at wind turbines}\label{sec:results}

\begin{figure}
\begin{center}

 \includegraphics[width=.75\textwidth]{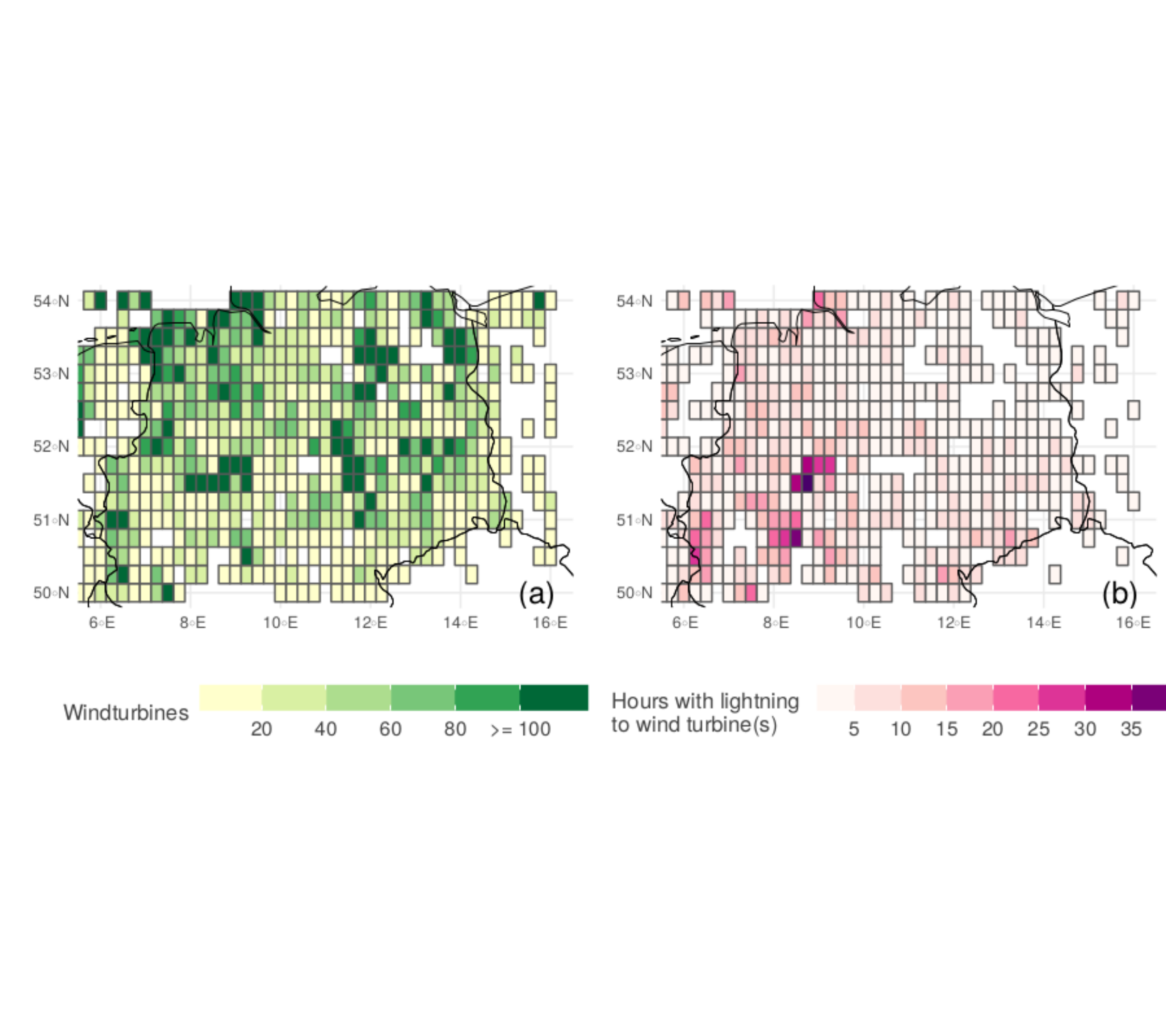}
  \caption{Panel (a): number of wind turbines per grid cell derived from © OpenStreetMap $2020$ data. Panel (b): number of hours per grid cell with lightning at wind turbines derived from EUCLID data. }
  \label{fig:gwt_gflashes}
 \end{center}
\end{figure}

The extraction of wind turbine locations and identification of lightning strikes to them within $300$~m in the colder season (ONDJFMA) shows that there are regions within the study domain that experience UL more frequently than others (see Fig.~\ref{fig:gwt_gflashes}). Interestingly, regions which are more often affected by UL (panel (b), dark pink) coincide with regions with many wind turbines. However, in general it can be observed that regions with a high number of wind turbines (panel (a), dark green) do not necessarily coincide with a high number of UL as can be seen in the North-Eastern parts of the study domain, for instance.

The following sections present and discuss the results when extending the findings from the instrumented towers to the study domain in which wind turbine locations are extracted and the lightning activity to them is analyzed.

\subsection{Diagnosing UL-LLS at wind turbines from larger-scale meteorological conditions}\label{sec:results1}

The random forest models for UL-LLS and UL-LLS + UL-noLLS based on data from the two instrumented towers identified larger-scale meteorological variables which are most important distinguishing situations with and without UL. 

Now, the tower-trained random forest models are applied to the larger study domain to assess the risk of UL at wind turbines. Lightning measurements from LLS data shall verify the results at identified wind turbine locations.  

%TODO hier ganz klein blitze einfuegen!
\begin{figure}
\begin{center}
 \includegraphics[width=0.8\textwidth]{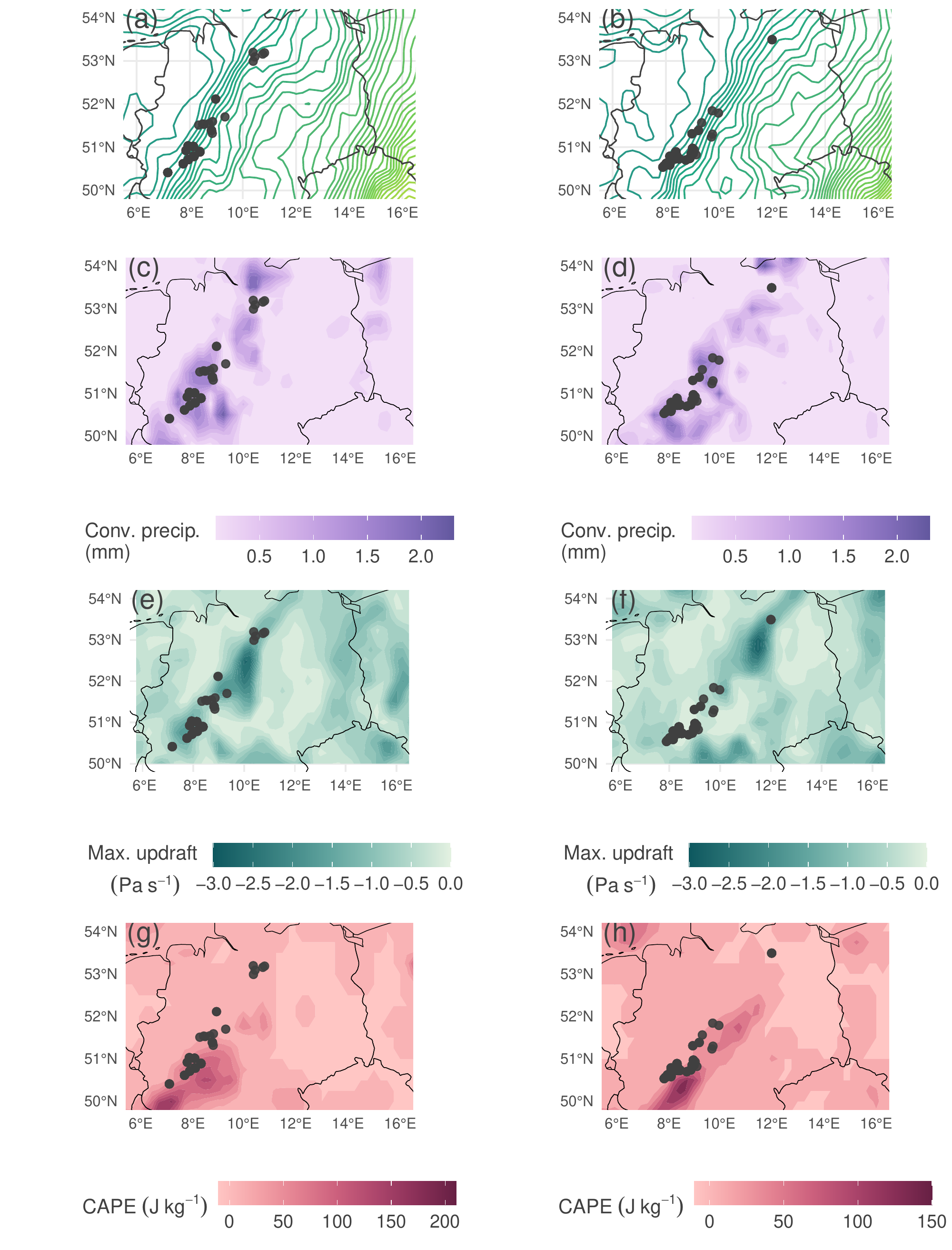}
  \caption{Larger-scale meteorological setting on the $4th$ March 2019 over the study domain. Left column illustrates the setting at 13 UTC, right column at 14 UTC. From upper to lower: spatial distributions of isolines of the $850$~hPa temperature (in intervals of 1~K), convective precipitation, the maximum large-scale updraft velocity (negative values is upward motion) and CAPE. Darker colors indicates higher magnitude. Dark gray dots in all figures are flashes within the considered hour and ERA5 grid cell derived from LLS EUCLID data.}
  \label{fig:meteo_case_2019_03_04}
 \end{center}
\end{figure}

%TODO pixel da wo keine wind turbinen sind anstelle alpha flaeche oder so
\begin{figure}
\begin{center}
 \includegraphics[width=.7\textwidth]{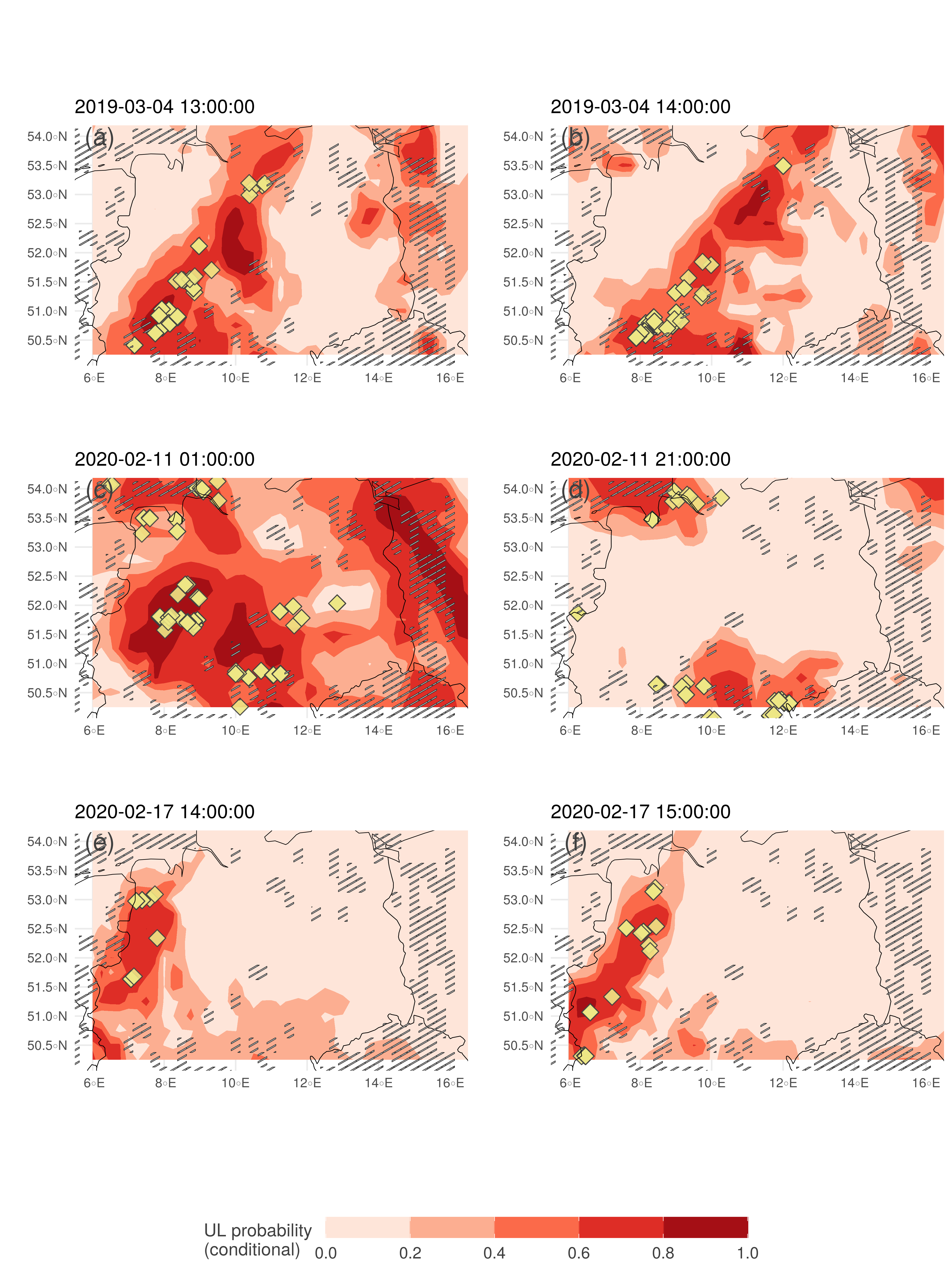}
  \caption{Median diagnosed conditional probability of UL according to $100$ random forest models based on Gaisberg and S\"antis Tower data (red areas). Yellow symbols are flashes within the considered hour derived from EUCLID data. Gray shaded areas are grid cells without wind turbines.}
  \label{fig:case_studies1}
 \end{center}
\end{figure}

The following results are based on a similar procedure as described in Sect.~\ref{sec:methods2} except that each grid cell ( 31 km x 31 km ) of the study domain is used as test data instead of the cross-validated data from the instrumented towers. 

In the following, the tower-trained random forest models are applied to each grid cell of the study domain.
% TODO: braucht es das?
To increase the robustness of the results, again $100$ different random forest models based on observations from the Gaisberg and the S\"antis Tower are used to diagnose the conditional probability of UL on the selected case studies over the study domain. The results in this section visualize the median conditional probabilities diagnosed by the random forest models.

\subsubsection*{Case studies: UL-LLS at wind turbines}\label{sec:casestudies}
 To illustrate the diagnostic ability of the tower-trained random forests for UL-LLS on days with UL events, three different case study days are selected out of colder seasons between $2018$ and $2020$ in the study domain.

The selected case study days are characterized by typical weather situations for the colder seasons in the mid-latitudes. The atmosphere in the transition seasons and in winter tends to be highly variable and influenced by the succession of cyclones and anticyclones determining the meteorological setting \citep{perry1987}.

In particular the development and progression of mid-latitude cyclones provides favorable conditions for so-called wind-field thunderstorms \citep{morgenstern2022}. This thunderstorm type is among others associated with strong updrafts, high amounts of precipitation as well as low but present CAPE. 
%A recent study shows that the dominant thunderstorm type producing UL at the Gaisberg and S\"antis Tower is the wind-field thunderstorm as well as wind-field thunderstorms with enhanced cloud physics such as large amounts of solid and liquid particle loadings (TODO:ICLP paper zitat??). %TODO:\citep{stucke2022b}. 

The first case study is considered in more detail regarding the drivers identified at the instrumented towers (Fig.~\ref{fig:varimp}). Figure~\ref{fig:meteo_case_2019_03_04} illustrates the larger scale isotherm locations, the spatial distribution of convective precipitation, the maximum updraft velocity and CAPE on the $4th$ March $2019$ at 13 UTC and 14 UTC. LLS detected lightning events to the identified wind turbines within the particular hour are indicated as dark gray dots.

The meteorological setting is determined by the passage of a cold front ahead of a trough around noon. Densely packed isotherms at $850$~hPa crossing Northern and Central Germany from West to East indicate the approximate location of the cold front in panels (a) and (b).
The cold front implies locally enhanced amounts of convective precipitation in (c) and (d), strong updrafts indicated by large negative values in (e) and (f) and slightly increased but in general low CAPE in (g) and (h) in comparison to deep convection in summer. All three variables show maximum increased values in slightly different areas within the study domain induced by the cold front. Convective precipitation shows increased values along the cold front, whereas the other two variables have locally more concentrated areas with maximum values (e.g., maximum updraft velocity in North/Central Germany). 

Figure~\ref{fig:case_studies1} visualizes the diagnosed conditional probability by the random forest models in red colors for all three case study days. Panels (a) and (b) show the results for the particular case study discussed in Fig.~\ref{fig:meteo_case_2019_03_04}. The diagnosed pattern is a result of combining the influence of the three driver variables. This suggests that not a single variable can be responsible for the resulting probability map but it is rather an interaction of different influential variables yielding areas with increased risk to experience UL. 

The yellow symbols again show lightning strikes over the considered hour. Identified lightning events in yellow require a wind turbine within a distance of maximum $300$~m as described in Sect.~\ref{sec:data}. All other tall structures that might have experienced UL are not considered in this figure.
Therefore, the diagnosed probabilities do not depend on wind turbine locations meaning that high probabilities might be diagnosed even though there is no wind turbine installed. Grid cells without any wind turbines are shaded in gray.

All three case study days in Fig.~\ref{fig:case_studies1} show that areas with increased diagnosed probability for UL to occur coincide well with identified lightning events in the respective hour over the study domain. In all three case studies there is a clear separation between areas with very low diagnosed risk and areas with very high diagnosed risk to experience UL.

On the $11th$ February 2020 shown in panel (c) and (d) of Fig.~\ref{fig:case_studies1}, the study domain is in strong westerly flow again associated with locally increased convective precipitation, CAPE and strong updrafts (not shown here).
On the $17th$ February 2020, the study domain is crossed by a cold front in higher altitudes (above $500$~hPa). Regardless of the different meteorological situation, the conditions are again similar to the other case studies showing increased values in the three driver variables that highly influence the diagnosed conditional probability.

\subsection{Risk assessment of UL at wind turbines}\label{sec:results2}

Identifying areas with increased risk of UL due to larger-scale meteorological conditions is a valuable step towards the risk assessment of lightning at wind turbines. 
The case studies clearly demonstrate that observed lightning at wind turbines coincide with areas of increased probability diagnosed by the random forest models. The following analysis considers all events within the considered period of time in which lightning at wind turbines was identified.
Not only the models for UL-LLS shall provide a risk assessment, but now random forests for UL-LLS + UL-noLLS are additionally applied to the study domain and the considered time period.

The considered study period including the transition seasons and winter from $2018$ to $2020$ counts in total $185$ event days with $1~027$ single flash hours and $18~602$ single flash events. These numbers shall be a measure to verify the resulting diagnosed probabilities by the random forest models. 
Note that these numbers are the lower limit of actually occurred flashes. Considering the uncertainty of manually identifying flashes at wind turbines as well as the uncertainty of detecting UL by the LLS suggests a significantly larger number of actually occurred lightning events at wind turbines. Further, this verification approach exclusively considers lightning at wind turbines and neglects all other tall structures such as radio towers in the study domain that might be affected by UL.
In the following, all days within the considered study period are taken as new data for the random forest models to diagnose the conditional probabilities on hourly basis. 

The objective is to identify regions that are more frequently affected by a higher risk of UL compared to other regions according to the random forest models. 
For this purpose the number of hours in each ERA5 grid cell (~0.25~°~x~0.25~°~) that exceeds the conditional probability threshold of $0.5$ is counted.

\subsubsection*{Risk assessment of UL-LLS at wind turbines}
Figure \ref{fig:probcounts_ICC} (a) illustrates that there are regions in the considered study domain having a higher risk to experience UL-LLS more often than other regions. The western and southwestern parts of the study domain have a considerably higher probability for UL-LLS.
This is also in agreement with panel (b) in Fig.~\ref{fig:gwt_gflashes} showing the actually observed hours in which at least one lightning event to a wind turbine occurred within the respective grid cell. 

%Interestingly parts of the area in Nordrhein-Westfalen showing the highest risk of UL coincide with areas with a very low number of wind turbines. This particular area also coincides with a slightly elevated topography. Comparing Fig.~\ref{fig:windmills_topomap} with Fig.~\ref{fig:probcounts_ICC} shows that topography might influence the diagnosed conditional probability of UL. This increased probability due to elevated topography does, however, often coincide with a low number of installed wind turbines explaining the lower number of observed lightning despite higher diagnosed risk. Even though the change in the topography in the study domain is rather small, the effect it has on the probability of UL is clearly visible. In this context it would be interesting to further investigate this effect in other regions such as regions with more complex terrain for instance.  
Interestingly, areas with higher UL-LLS probabilities in Fig.~\ref{fig:probcounts_ICC} roughly coincide with regions of elevated topography in the southern third of the domain (cf. Fig.~\ref{fig:windmills_topomap}). Possible explanations are an increased lightning-effective height \citep[e.g.,][]{shindo2018lightning} of the turbines and increased chances for thunderstorm formation through orographic lifting and thermally-induced breezes \citep{kirshbaum2018}. Sea breezes might also be an explanation for the higher probabilities in the northwesternmost, sea-covered part of the domain.

The successful transfer of the UL-LLS model trained with meteorological data on direct tower measurements to a larger region and its independent verification on wind turbines shows the potential of our approach to be able to produce regionally varying risk maps, which might in turn lead to regionally varying (voluntary or enforced) lightning protection standards for wind turbines. 

%Observations at the Gaisberg Tower reveal that more than $50$~\% of UL is not detected by EUCLID.
%Extending the analysis using random forest models based on all subtypes of UL observed at the Gaisberg Tower might give a more realistic order of magnitude of UL affecting the study domain. 
\subsubsection*{Risk assessment of UL-LLS + UL-noLLS at wind turbines}

The successful transfer of the tower-trained and verified UL-LLS model to a larger domain lends credence to taking the same step with the tower-trained model for all upward lightning (UL-LLS and UL-noLLS) although no data exist for an independent verification.

\begin{figure}
\begin{center}

 \includegraphics[width=.9\textwidth]{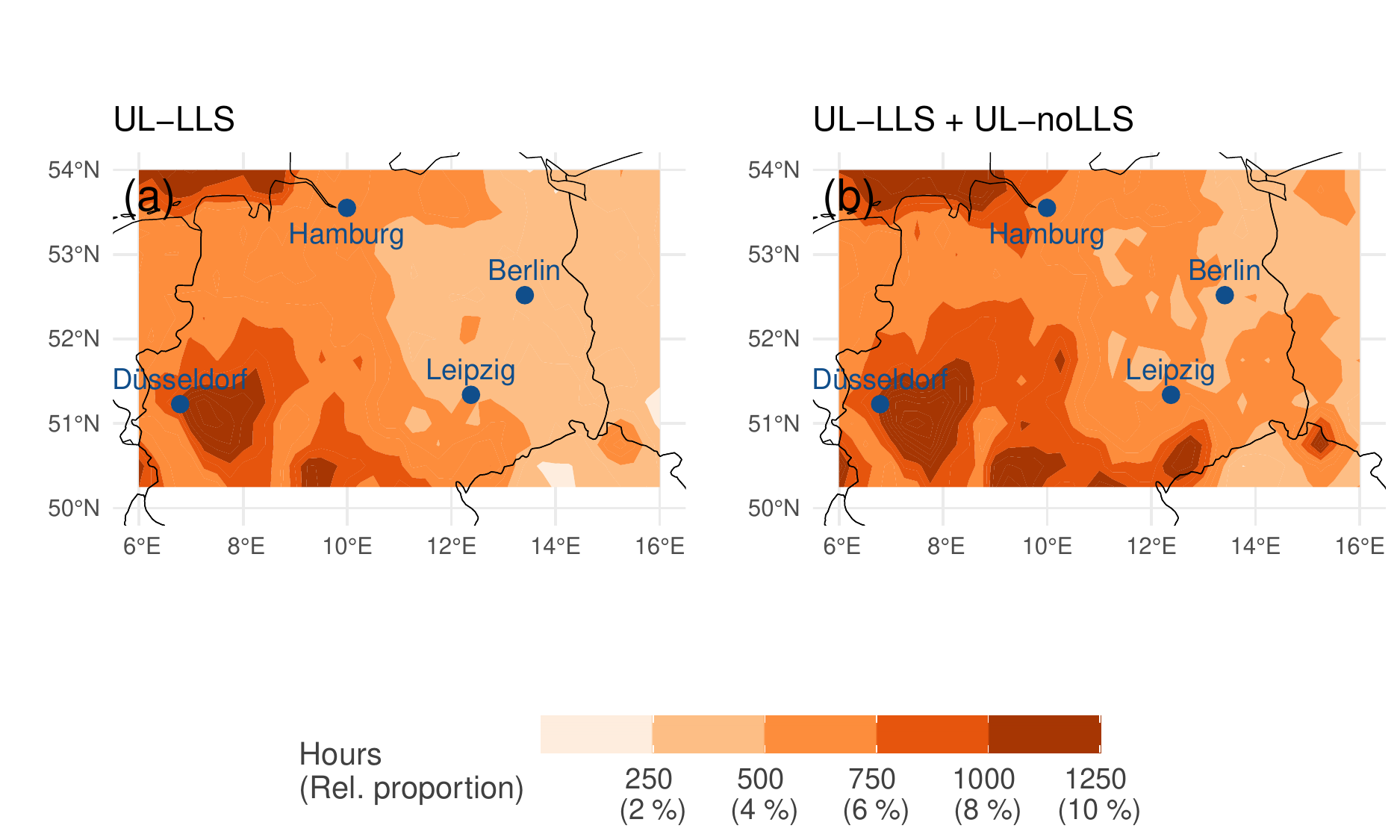}
  \caption{Panels (a) and (b): potential maps for UL in the colder season (ONDJFMA) from $2018$ to $2020$. Orange colors are median of hours per grid cell exceeding conditional probabilities of $0.5$ according to $100$ random forest models. Panel (a) shows results according to models based on Gaisberg and S\"antis data combined. Panel (b) shows results according to models based on Gaisberg data also including the UL-noLLS. Relative proportion of in total 12480 hours are given as reference.}
  \label{fig:probcounts_ICC}
 \end{center}
\end{figure}

Panel (b) in Fig.~\ref{fig:probcounts_ICC} indicates that more flashes are expected when additionally accounting for the LLS-non detectable UL flash type. The pattern of areas with increased risk to experience UL are similar even though some areas affected more often are enlarged. From this it can be suggested that there are similar mechanisms that result from larger-scale meteorology leading to the UL-LLS or UL-noLLS flash types. %It is rather the order of magnitude of UL events that is different when additionally accounting for ICC\textsubscript{only} UL flashes.

The risk in regions with elevated topography in the southern part of the domain and in the coastal northwesternmost region is most pronouncedly increased.

%TODO ueber balken prozentzahl: 12480 = 100 %
                       %         250 = 2 %
                        %        750 = 6 %
                         %       1000 = 8
                          %      1250 = 10
                          
                      %    100 0.8
                      %    200 1.6
                      %    300 2.4
                      %    400 3.2
                      %    500 4
                      %    600 4.8

%\begin{figure}
%\begin{center}

% \includegraphics[width=.75\textwidth]{probcounts_ICC}
%  \caption{Panels (a)--(d): potential maps for UL based on $185$ days in the colder season (ONDJFMA) from $2018$ to $2020$. Orange colors are median of hours per grid cell exceeding conditional probabilties of $0.5$ (left column) or $0.8$ (right column) according to $100$ random forest models. Panels (a) and (b) show results according to models based on Gaisberg and S\"antis data combined. Panels (c) and (d) show results according to models based on Gaisberg data only including the LLS-non-detectable subtype of UL. Panel (e) shows number of wind turbines per grid cell derived from OpenStreetMap data. Panel (f) shows the number of hours per grid cell with lightning at wind turbines derived from EUCLID data. }
%  \label{fig:probcounts_ICC}
% \end{center}
%\end{figure}

%reduced version

\section{Conclusions}\label{sec:conclusions}

Upward lightning (UL) initiating at tall structures such as wind turbines is much more destructive than  downward lightning (DL). Each UL flash starts with an initial continuous current (ICC) lasting about ten times longer than in DL transferring much more charge to the tall structure. Further, direct upward lightning measurements suggest that less than $50$~\% of UL events can be detected by most lightning location systems (LLS) since they are not able to spot UL with only an ICC. 

UL directly measured at the instrumented tower at Gaisberg has little seasonal variation. However, current lightning protection standards are based on the annual flash density derived from LLS data which is clearly dominated by DL in the warm season. UL-noLLS is completely neglected and UL in the cold season is highly underestimated.
Basic knowledge about the occurrence of UL is still incomplete impeding a proper risk assessment of UL at wind turbines. 

The missing consideration of UL-noLLS and of the importance of the cold season for UL will therefore considerably underestimate the risk of UL to wind turbines. This study leverages rare direct UL measurements with larger-scale meteorological data in a machine learning model in order to estimate the risk of all UL (UL-LLS and UL-noLLS) at wind turbines.

The first step constitutes training and validating two different random forest models based on long-term observations from two specially instrumented towers. One model accounts only for UL-LLS and one model accounts for UL-LLS + UL-noLLS.
The model input data are direct UL measurements from the Gaisberg Tower (Austria, $2000$-$2015$) and from the S\"antis Tower (Switzerland, $2010$-$2017$). While the sensor at the Gaisberg Tower measures also UL-noLLS, the sensor at the S\"antis Tower misses most of them.

In a second step, the random forest models are extended to a larger study domain (50°N - 54°N and 6°E -16°E) to identify areas with increased risk of UL in the colder season (ONDJFMA). As a verification, all lightning strikes at wind turbines in this domain are extracted from LLS and OpenStreetMap data and compared to the diagnosed probabilities by the random forests. 

Results show that UL can be reliably diagnosed by the tower-trained random forest models at the Gaisberg and S\"antis Tower. The larger-scale meteorological drivers are large amounts of (convective) precipitation, strong vertical updraft velocities and slightly increased CAPE. Further, the vertical extent of the cloud as well as the amount of ice crystals and solid hydrometeors are important variables.

The extension of the random forests to a larger domain shows that probability maps coincide with observed lightning strikes at wind turbines. Extending models trained at the Gaisberg Tower including UL-noLLS flashes reveals that areas with increased risk to experience UL are expected to experience UL even more often.

The western and southern part of the domain in North-West Germany with elevated topography and the coastal region in its northwesternmost part are most at risk of UL at wind turbines. 
This study demonstrates that direct UL measurements at an instrumented tower can be reliably modeled from larger-scale meteorological conditions in a machine learning model (random forest). The study also proposes a novel way how the transfer of that model to a larger region can be justified by using UL-LLS data at wind turbine locations. Consequently regionally detailed risk maps of UL at wind turbines can be produced. 

\clearpage
\noappendix       %% use this to mark the end of the appendix section. Otherwise the figures might be numbered incorrectly (e.g. 10 instead of 1).

%\appendixfigures
\appendix
\section{Additional material}\label{sec:appendix}
\subsection{Data availability}
ERA5 data are freely available for download at https://cds.climate.copernicus.eu
\cite{hersbach2020}. EUCLID data and direct observations from the Gaisberg Tower are available only on request. For more details contact Wolfgang Schulz.

\subsection[Software]{Software}\label{sec:app_software}    %% Appendix A1
All calculations as well as setting up the final data sets, modeling and predicting were performed using R \citep{R}, using packages netCDF4 \citep{pkgncdf4},  partykit  \citep{hothorn2015}, ggplot2 package \citep{pkgggplot2}. Retrieving the raw data and deriving further variables from ERA5 required using Python3 \citep{python3} and cdo \citep{cdo}.

\subsection{Risk assessment of UL at wind turbines using a higher probability threshold}

\begin{figure}
\begin{center}

 \includegraphics[width=.9\textwidth]{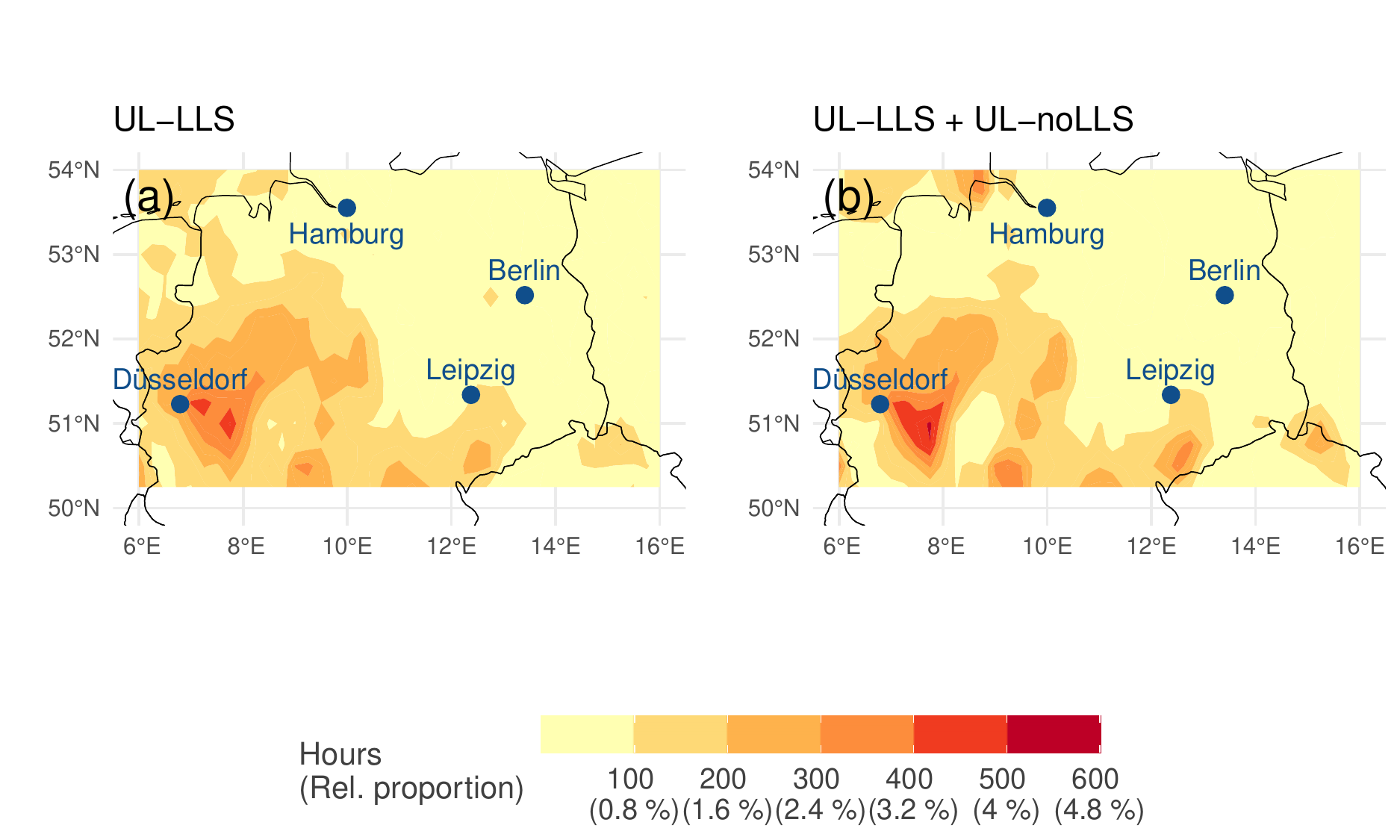}
  \caption{Panels (a) and (b): maps for the potential of UL in the colder season (ONDJFMA) from $2018$ to $2020$. Orange colors are median of hours per grid cell exceeding conditional probabilities of $0.8$ according to $100$ random forest models. Panel (a) shows results according to models based on Gaisberg and S\"antis data combined. Panel (b) shows results according to models based on Gaisberg data also including the UL-noLLS. Relative proportions of in total 12480 hours are given as reference.}
  \label{fig:probcounts_ICC_app}
 \end{center}
\end{figure}

In Sect.~\ref{sec:results2} the model results for the risk assessment of UL-LLS and UL-LLS + UL-noLLS are presented in the way that hours are counted exceeding a conditional probability of $0.5$.
Figure~\ref{fig:probcounts_ICC_app} illustrates the risk assessment using a higher probability threshold, namely $0.8$. The number of hours exceeding this threshold is lower by about a factor of two in comparison to a probability threshold of $0.5$. However, the regional pattern is still similar with maxima West/South-West of the study domain. 

\appendixfigures  %% needs to be added in front of appendix figures

\clearpage
%\subsection{Variables and variable groups}\label{sec:app_vars}
\appendixtables
\begin{table}
\newcommand{\tabincell}[2]{\begin{tabular}{@{}#1@{}}#2\end{tabular}}
	\caption{Table of large-scale variables taken from ERA5 and variables derived from ERA5. The derived variables (indicated in italics) are suggested to be potentially important in the charging process of a thundercloud or for the development of convection.}
	\setlength\extrarowheight{-6pt}
	\begin{tabular}{ll}
		\middlehline
		\textbf{Large-scale variables} & \textbf{Unit} \\
		\middlehline
	              	\specialrule{0em}{6pt}{6pt} 
		               cloud base height above ground  & m agl\\
		              \specialrule{0em}{4pt}{4pt}
		               \tabincell{l}{convective precipitation \\ (rain + snow)} & m\\
		              \specialrule{0em}{4pt}{4pt}
		               large scale precipitation & m \\
		               \specialrule{0em}{4pt}{4pt}
		              cloud size & m\\
		              \specialrule{0em}{4pt}{4pt}
		               \tabincell{l}{maximum precipitation rate \\ (rain + snow)} & kg~m$^{-2}$~s$^{-1}$ \\
		              \specialrule{0em}{4pt}{4pt}
		               ice crystals (total column, tciw) & kg~m$^{-2}$\\
		              \specialrule{0em}{4pt}{4pt}
		               Solid hydrometeors (total column, tcsw) & kg~m$^{-2}$\\
		              \specialrule{0em}{4pt}{4pt}
		               \tabincell{l}{supercooled liquid water \\(total column, tcslw)} & kg~m$^{-2}$ \\
		              \specialrule{0em}{4pt}{4pt}
		               water vapor (total column) & kg~m$^{-2}$\\
		              \specialrule{0em}{4pt}{4pt}
		               \tabincell{l}{vertical integral of divergence \\ of cloud frozen water flux} & kg~m$^{-2}$~s$^{-1}$ \\
		              \specialrule{0em}{4pt}{4pt}
		               \tabincell{l}{\emph{vertical transport of liquids} \\ \emph{around $-10$ °C}} & kg~Pa~s$^{-1}$\\
		              \specialrule{0em}{4pt}{4pt}
		               \tabincell{l}{\emph{ice crystals} \\  \emph{($-10$ °C - $-20$ °C)}} & kg~m$^{-2}$\\
		              \specialrule{0em}{4pt}{4pt}
		               \tabincell{l}{\emph{ice crystals} \\  \emph{($-20$ °C - $-40$ °C)}} & kg~m$^{-2}$\\
		              \specialrule{0em}{4pt}{4pt}

		             \tabincell{l}{\emph{cloud water droplets} \\ \emph{($-10$ °C - $-20$ °C)}} & kg~m$^{-2}$\\
		              \specialrule{0em}{4pt}{4pt}
		               \tabincell{l}{\emph{solid hydrometeors} \\  \emph{($-10$ °C - $-20$ °C)}} & kg~m$^{-2}$\\
		              \specialrule{0em}{4pt}{4pt}
		               \tabincell{l}{\emph{solid hydrometeors} \\  \emph{($-20$ °C - $-40$ °C)}} & kg~m$^{-2}$\\
		              \specialrule{0em}{4pt}{4pt}
		               \tabincell{l}{\emph{solids (cswc + ciwc)} \\  \emph{around $-10$ °C}} & kg~m$^{-2}$ \\
		              \specialrule{0em}{4pt}{4pt}
		               \tabincell{l}{\emph{liquids (clwc + crwc)} \\  \emph{around $-10$ °C}} & kg~m$^{-2}$\\
		              \specialrule{0em}{4pt}{4pt}
			              2 m dew point temperature & K\\
 		              	  	\end{tabular}
 %	\label{tab:era5_names}
 \end{table}
 \begin{table}
 \newcommand{\tabincell}[2]{\begin{tabular}{@{}#1@{}}#2\end{tabular}}
 	\setlength\extrarowheight{-6pt}
 	\begin{tabular}{ll}
% 	\middlehline
 %	\specialrule{0em}{4pt}{4pt}
 	                
 	              %  \middlehline
% 	                \specialrule{0em}{6pt}{6pt}
                   \specialrule{0em}{4pt}{4pt}
		                \tabincell{l}{mean vertically integrated \\ moisture convergence} & kg~m$^{-2}$~s$^{-1}$\\
		              
		               \specialrule{0em}{4pt}{4pt}

		                \tabincell{l}{\emph{water vapor} \\ \emph{($-10$ °C - $-20$ °C)}}  & kg~m$^{-2}$ \\
		              
		               \specialrule{0em}{4pt}{4pt}

		                  boundary layer height& m \\
	                  	\specialrule{0em}{4pt}{4pt}

		                   surface latent heat flux  & J~m$^{-2}$\\
		                
		                  \specialrule{0em}{4pt}{4pt}
		                   surface sensible heat flux  & J~m$^{-2}$\\
		                  \specialrule{0em}{4pt}{4pt}
		                   downward surface solar radiation  & J~m$^{-2}$\\
		                  \specialrule{0em}{4pt}{4pt}

		                   \tabincell{l}{convective available \\  potential energy}  & J~kg$^{-1}$\\
		                
		                  \specialrule{0em}{4pt}{4pt}
		                   convective inhibition present & binary\\
		                   \specialrule{0em}{4pt}{4pt}
		                   mean sea level pressure & Pa\\

		                  \specialrule{0em}{4pt}{4pt}
		                   \emph{height of $-10$ °C isotherm} & m agl\\
		                \specialrule{0em}{4pt}{4pt}
		                   boundary layer dissipation  & J~m$^{-2}$\\
                      \specialrule{0em}{4pt}{4pt}
		                   \emph{Maximum vertical updraft velocity}  & Pa~s$^{-1}$ \\
		                  \specialrule{0em}{4pt}{4pt}
		                   \emph{total cloud shear} & m~s$^{-1}$ \\
                        \specialrule{0em}{4pt}{4pt}
		                   \emph{wind speed at 10 m} & m~s$^{-1}$\\
		                  \specialrule{0em}{4pt}{4pt}
		                   \emph{wind direction at 10 m} & $^\circ$\\
		                 
		                  \specialrule{0em}{4pt}{4pt}
		                   \tabincell{l}{\emph{shear between 10 m and cloud base}} & m~s$^{-1}$\\
		                  \specialrule{0em}{4pt}{4pt}
		                 %  \middlehline   
		                  
\bottomhline
	\end{tabular}
	\label{tab:era5_names}
\end{table}

\clearpage

\begin{acknowledgements}

We acknowledge the funding of this work by the Austrian Research Promotion
Agency (FFG), project no.~872656 and Austrian Science Fund (FWF) grant
no.~P\,31836.
We thank the EMC Group of the Swiss Federal Institute of Technology (EPFL) for providing the data of the S\"antis Tower strikes.
Finally we thank Siemens, the operator of BLIDS for providing EUCLID data.
\end{acknowledgements}

\authorcontribution{Isabell Stucke did the investigation, wrote software,
visualized the results and wrote the paper. Deborah Morgernstern, Thorsten Simon and
Isabell Stucke performed data curation, built the data set, and derived
variables based on ERA5 data. Thorsten Simon contributed with coding concepts.
Georg J. Mayr provided support on the meteorological analysis, data organization
and funding acquisition. Achim Zeileis supervised the formal analysis and
interpretation of the statistical methods. Achim Zeileis, Georg J. Mayr, and
Thorsten Simon are the project administrators and supervisors. All authors
contributed to the conceptualization of this paper, discussed on the
methodology, evaluated the results, and commented on the paper.}

\competinginterests{The authors declare that they have no conflict of interest.}

\bibliographystyle{copernicus}
\bibliography{references.bib}

\end{document}